\documentclass{article}

\usepackage{arxiv}

\usepackage[utf8]{inputenc} 
\usepackage[T1]{fontenc}    
\usepackage{hyperref}       
\usepackage{url}            
\usepackage{booktabs}       
\usepackage{amsfonts}       
\usepackage{nicefrac}       
\usepackage{microtype}      
\usepackage{lipsum}		    
\usepackage{graphicx}
\usepackage{doi}

\usepackage[table]{xcolor}  
\usepackage{colortbl}       
\usepackage{booktabs}       
\usepackage{graphicx}       
\usepackage[numbers]{natbib}    
\usepackage{amsmath} 
\usepackage{float} 
\usepackage{glossaries}
\makeglossaries
\newacronym{FNAB}{FNAB}{Fine Needle Aspiration Biopsy}
\newacronym{AUC}{AUC}{Area Under the Curve}
\newacronym{CNN}{CNN}{Convolutional Neural Network}
\newacronym{ROI}{ROI}{Region of Interest}
\newacronym{Grad-CAM}{Grad-CAM}{Gradient-weighted Class Activation Mapping}
\newacronym{TBSRTC}{TBSRTC}{The Bethesda System for Reporting Thyroid Cytopathology}
\newacronym{PTC}{PTC}{Papillary Thyroid Carcinoma}
\newacronym{WSI}{WSI}{Whole Slide Image}
\newacronym{CODT}{CODT}{Correlation Optical Diffraction Tomography}
\newacronym{IOPD}{IOPD}{Intraoperative Thyroid Nodule Diagnosis}
\newacronym{SVM}{SVM}{Support Vector Machine}
\newacronym{FFNN}{FFNN}{Feed-Forward Neural Network}
\newacronym{PCA}{PCA}{Principal Component Analysis}
\newacronym{t-SNE}{t-SNE}{t-distributed Stochastic Neighbor Embedding}
\newacronym{UMAP}{UMAP}{Uniform Manifold Approximation and Projection}
\newacronym{BENIGN}{Benign}{Bethesda II: Observation / Non-Surgical Management}
\newacronym{INDET_SUS}{Indeterminate/Suspicious}{Bethesda I, III, IV, V: Further Investigation / Consideration for Intervention}
\newacronym{MALIGNANT}{Malignant}{Bethesda VI: Surgical Intervention}

\title{ThyroidEffi 1.0: A Cost-Effective System for High-Performance Multi-Class Thyroid Carcinoma Classification}

\author{
  \href{https://orcid.org/0009-0001-6723-0893} {\includegraphics[scale=0.06]{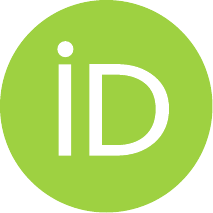}\hspace{1mm} Hai Pham-Ngoc} \\
  Data Science Laboratory \\ 
  Vietnam National University \\
  334 Nguyen Trai, Thanh Xuan, Hanoi \\
  \texttt{harito.work@gmail.com} \\
  \And
  De Nguyen-Van \\
  Department of Pathology (C2-E) \\ 
  108 Military Central Hospital \\
  1 Tran Hung Dao, Hai Ba Trung, Hanoi \\
  \texttt{doctorde108@gmail.com} \\
  \And
  Dung Vu-Tien \\
  Data Science Laboratory \\ 
  Vietnam National University \\
  334 Nguyen Trai, Thanh Xuan, Hanoi \\
  \texttt{duzngvt@gmail.com} \\
  \And
  Phuong Le-Hong\thanks{Corresponding author} \\
  Data Science Laboratory \\ 
  Vietnam National University \\
  334 Nguyen Trai, Thanh Xuan, Hanoi \\
  \texttt{phuonglh@vnu.edu.vn} \\
}

\date{}
\renewcommand{\shorttitle}{}

\begin{document}
\maketitle

\begin{abstract}

\textbf{Background}: Automated classification of thyroid \gls{FNAB} images faces challenges in limited data, inter-observer variability, and computational cost. Efficient, interpretable models are crucial for clinical support.
\textbf{Objective}: To develop and externally validate a deep learning system for the multi-class classification of thyroid \gls{FNAB} images into three key categories that directly guide post-biopsy treatment decisions in Vietnam: \gls{BENIGN}, \gls{INDET_SUS}, and \gls{MALIGNANT}, while achieving high diagnostic accuracy with low computational overhead.
\textbf{Methods}: Our pipeline features: (1) YOLOv10-based cell cluster detection for informative sub-region extraction and noise reduction; (2) a curriculum learning-inspired protocol sequencing localized crops to full images for multi-scale feature capture; (3) adaptive lightweight EfficientNetB0 (4 millions parameters) selection balancing performance and efficiency; and (4) a Transformer-inspired module for multi-scale, multi-region analysis. External validation used 1,015 independent \gls{FNAB} images.
\textbf{Results}: ThyroidEffi Basic achieved a macro F1 of 89.19\% and 
\gls{AUC}s of 0.98 (\gls{BENIGN}), 0.95 (\gls{INDET_SUS}), and 0.96 (\gls{MALIGNANT}) on the internal test set. External validation yielded \gls{AUC}s of 0.9495 (\gls{BENIGN}), 0.7436 (\gls{INDET_SUS}), and 0.8396 (\gls{MALIGNANT}). ThyroidEffi Premium improved macro F1 to 89.77\%. \gls{Grad-CAM} highlighted key diagnostic regions, confirming interpretability. The system processed 1000 cases in 30 seconds, demonstrating feasibility on widely accessible hardware like a 12-core CPU.
\textbf{Conclusions}: This work demonstrates that high-accuracy, interpretable thyroid \gls{FNAB} image classification is achievable with minimal computational demands.

\end{abstract}


\section*{Graphical Abstract}
\begin{figure}[H]
    \centering 
    \includegraphics[width=0.8\textwidth]{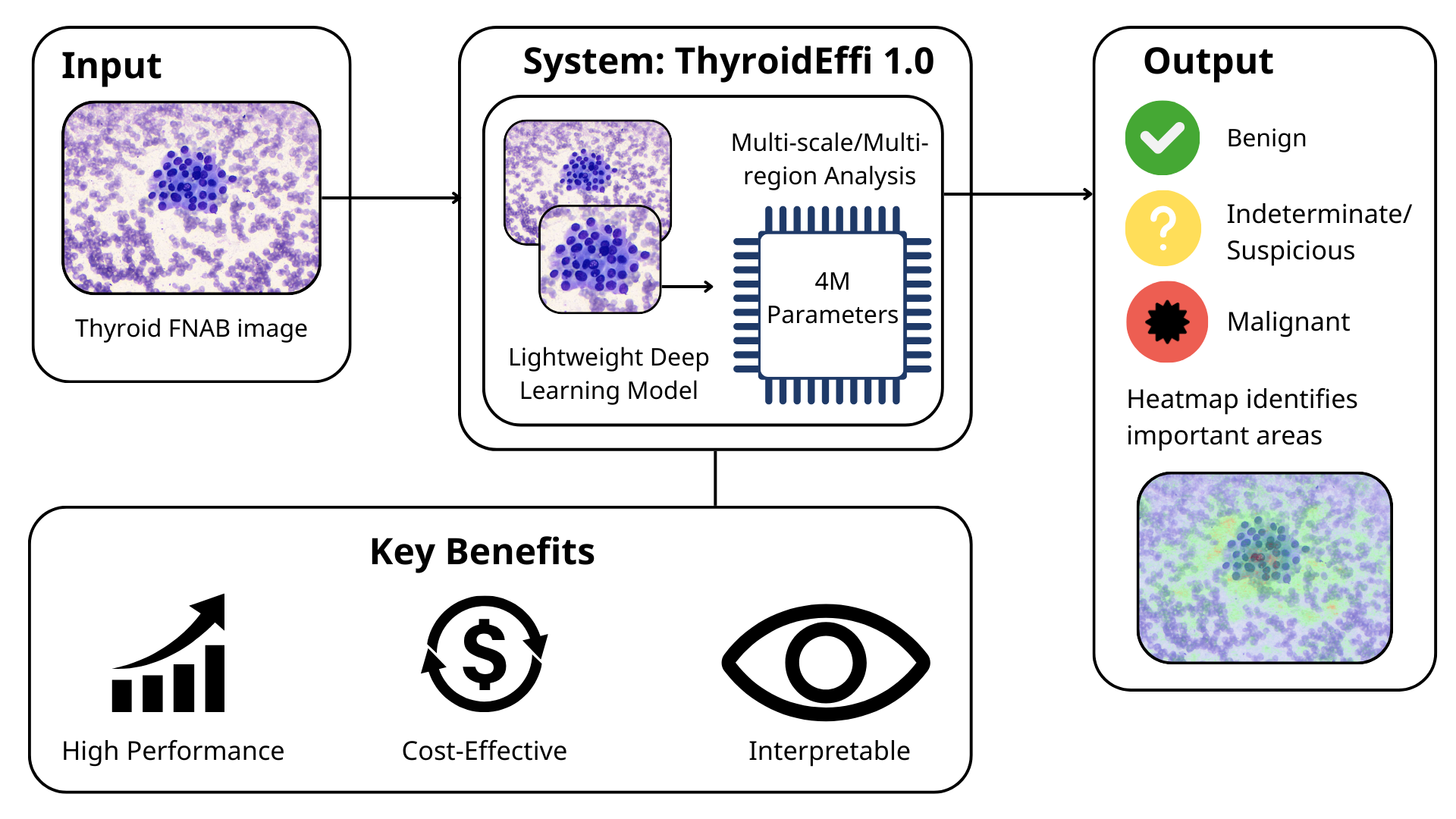} 
    \caption{Graphical Abstract}
    \label{fig:graphical_abstract}
\end{figure}

\section{Introduction}

\subsection{Background and Significance}

Thyroid carcinoma is a prevalent endocrine malignancy worldwide \cite{intro1}. Accurate and efficient diagnosis of thyroid conditions is crucial for effective treatment and improved patient outcomes. \gls{FNAB} is a minimally invasive procedure widely used for assessing thyroid abnormalities and guiding treatment decisions \cite{intro2}. Historically, \gls{FNAB} diagnosis has relied heavily on cytopathologists' expertise. However, this reliance leads to potential inter-observer variability \cite{intro3}. The global shortage of trained cytopathologists further emphasizes the need for automated diagnostic methods. Deep learning has shown promise in various medical imaging tasks \cite{intro4}. Moreover, cost-efficiency is essential for real-world deployment, particularly in resource-constrained settings.

\subsection{Literature Review}

Several studies have explored automated classification of thyroid cytology images using deep learning, employing various approaches. Early works often focused on binary classification using 
\gls{CNN}, such as \citet{or1_image}, which classified \gls{PTC} and non-\gls{PTC}. These studies demonstrated the potential of deep learning but oversimplified the diagnostic task by ignoring the nuanced categories of \gls{TBSRTC}. Other studies adopted a \gls{WSI} approach, utilizing two-stage \gls{CNN}s for \gls{ROI} identification and subsequent classification, as seen in \citet{or2_wsi}, \citet{or4_wsi}, \citet{or6_wsi}, \citet{or8_image}, and \citet{or9_wsi}. While \gls{WSI} analysis allows for comprehensive slide evaluation, it often suffers from increased computational complexity and requires specialized equipment. Another line of research focused on fragment-level analysis, as exemplified by \citet{or3_image} and \citet{or5_image}. These studies achieved high accuracy on fragmented images using models like VGG-16, Inception-v3, and DenseNet161, sometimes combined with techniques like AdaBoost. However, they faced challenges in aggregating fragment-level predictions to obtain reliable patient-level diagnoses and often required manual pre-processing. Some studies explored alternative techniques, such as \gls{CODT} combined with machine learning algorithms (\citet{or11_image}) or fluorescence polarization imaging (\citet{or14_image}) for cell segmentation and classification. While innovative, these approaches often require specialized equipment and may not be readily applicable in all clinical settings. More recently, studies have leveraged larger datasets and more sophisticated architectures, such as the ThyroPower system developed by \citet{or13_wsi}, demonstrating promising results in distinguishing benign from \gls{TBSRTC} III+ and V+ categories. \citet{or10_wsi} focused on \gls{IOPD} using surgical samples, combining computer vision techniques with \gls{CNN} and \gls{SVM}.

\subsection{Limitations of Existing Techniques}

Despite the extensive research, achieving robust multi-class classification of thyroid cytology images, aligned with \gls{TBSRTC}, remains a significant challenge \cite{review1, review2, review3, review4, review5, review6, review7, review8, review9, review10, review11, review12}.

\begin{figure}
    \centering
    \includegraphics[width=1\linewidth]{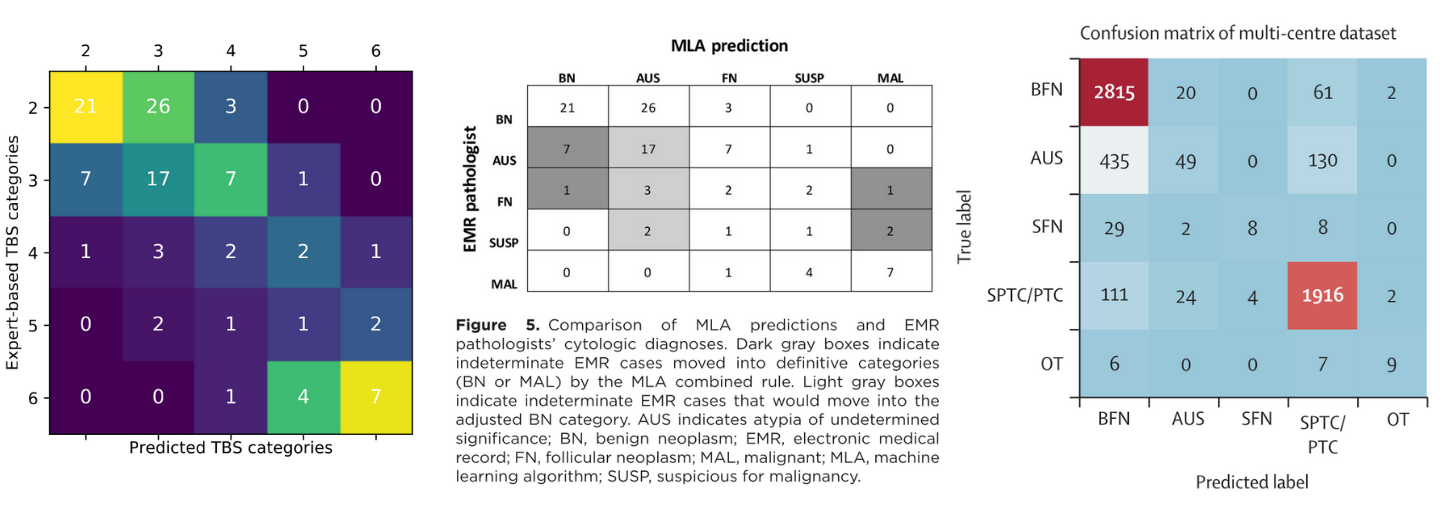}
    \caption{Test results from the limited number of studies classifying more than two classes are presented. From left to right: the first \cite{or2_wsi} and second \cite{or7_wsi} are from the United States research group, and the third \cite{or13_wsi} is from the China research group.}
    \label{fig1}
\end{figure}

Firstly, many are constrained by small datasets, hindering model generalizability. Secondly, patient-level prediction aggregation remains a non-trivial issue for fragment-based approaches. Thirdly, computational complexity often restricts the real-time applicability of intricate models, particularly those employing \gls{WSI} or multi-stage processing. Crucially, a key limitation lies in attaining consistently high multi-class classification performance. While recent investigations (from two research groups, one in China \citet{or13_wsi} and one in the United States \cite{or2_wsi, or7_wsi}) have explored multi-class classification, achieving consistently high accuracy across all categories, notably for categories V (Suspicious for Malignancy) and VI (Malignant), continues to be a major hurdle. Details are provided in Figure \ref{fig1}.

\subsection{Research Objectives}

\begin{figure}
    \centering
    \includegraphics[width=1\linewidth]{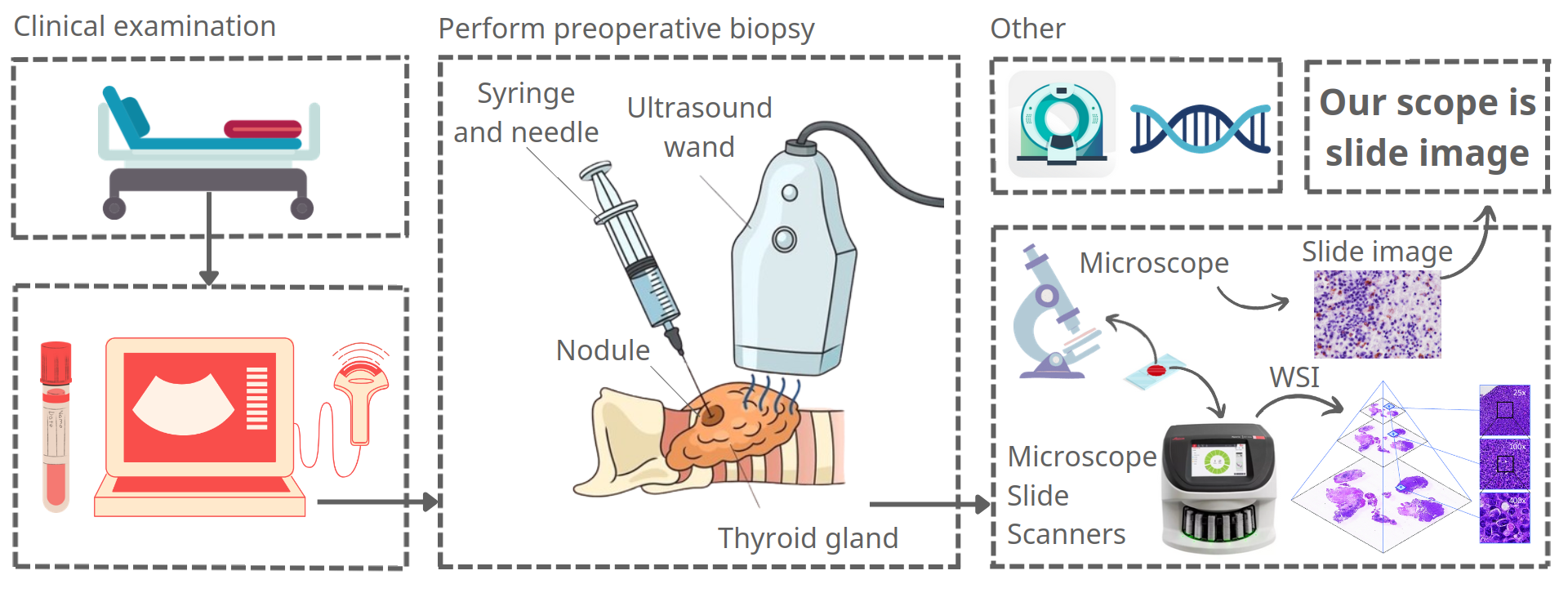}
    \caption{Workflow of thyroid nodule diagnosis, from clinical examination and ultrasound-guided biopsy to slide image analysis (our research scope).}
    \label{fig2}
\end{figure}

This research aims to develop a robust and cost-efficient deep learning model for the classification of thyroid nodule \gls{FNAB} slide images (the scope of this research is detailed in Figure \ref{fig2}). Our specific objectives are to achieve high classification accuracy; maintain low deployment cost (few parameters, low hardware requirements, fast inference time), which is particularly useful for practical implementations in resource-constrained settings like developing countries; and provide good interpretability of diagnostic decisions, enhancing clinical applicability.

\gls{TBSRTC} classifies thyroid nodule \gls{FNAB} results into six distinct categories (Bethesda I-VI). However, in clinical practice (especially in Vietnam), the interpretation of these six cytological categories ultimately guides patient management into one of three primary treatment pathways: \textbf{Observation/Non-Surgical Management} (typically corresponding to Bethesda II), \textbf{Further Investigation/Consideration for Intervention} (encompassing Bethesda I, III, IV, and V), or \textbf{Surgical Intervention} (primarily corresponding to Bethesda VI).

Our model is specifically designed to classify \gls{FNAB} images directly into these three clinically relevant groups: \gls{BENIGN} (corresponding to Bethesda II), \gls{INDET_SUS} (corresponding to the combined Bethesda I, III, IV, and V categories), and \gls{MALIGNANT} (corresponding to Bethesda VI). This approach is motivated by the direct link between these three groups and the most crucial decisions in patient care. Accurate classification into these groups provides crucial support for clinicians in determining the appropriate primary treatment pathway for a patient, streamlining the diagnostic process.

Compared to previous studies, which often focused solely on binary classification between benign and malignant, thus overlooking the crucial indeterminate and suspicious categories, or which addressed multi-class classification based on \gls{TBSRTC} but did not cover all six categories or evaluate performance comprehensively across all included classes, our research offers a more clinically aligned classification strategy. By training our model to classify directly into the three main management groups that encompass the entire Bethesda spectrum, this study provides a more pragmatic solution directly supporting the established clinical workflow, particularly beneficial for guiding treatment decisions based on current image data in resource-constrained settings.


\section{Dataset and Methods}

\subsection{Dataset}

The dataset consists of 1,804 high-resolution microscopic images obtained from \gls{FNAB} samples, each corresponding to a unique patient. These images were acquired at the 108 Military Central Hospital between 2018 and July 2024 using an Olympus BX43 optical microscope equipped with a high-resolution digital camera and a specific Diff-Quick staining protocol. The exclusive use of a 40x magnification and a standardized Diff-Quick staining protocol aimed to minimize inter-institutional variability in image acquisition. The dataset includes 482 (\gls{BENIGN}), 541 (\gls{INDET_SUS}), and 871 (\gls{MALIGNANT}) samples, reflecting the real-world distribution often encountered in clinical settings, with Bethesda VI (Malignant) cases being more prevalent. The selection of these three classes (\gls{BENIGN}, \gls{INDET_SUS}, \gls{MALIGNANT}) was based on their clinical significance and diagnostic challenge, as \textbf{accurately distinguishing between these groups is paramount for directing patients towards appropriate management pathways, ranging from observation to immediate surgical intervention}.

To ensure reproducibility and prevent data leakage, the dataset was split using a fixed random seed (42) with a 70:15:15 ratio into training, validation, and test sets. This split was performed \textbf{before} any data augmentation or model training. In other words, the operations described in \ref{method_m1}{ M1} – including the training of the YOLO model to automate cell cluster detection (defined as $\geq$ 10 cells) – were exclusively conducted on a pre-allocated subset of 120 manually annotated training images. Similarly, data augmentation was applied solely to the training set. The internal validation, test, and external validation sets remained completely isolated from any modifications or manipulations described in \ref{method_m1}{ M1}, \ref{method_m2}{ M2}, or any other process beyond their intended use for preventing overfitting (internal validation) and evaluating the trained model's performance (test and external validation). This separation is explicitly stated to avoid any misunderstanding regarding potential data leakage or manual curation bias.

\begin{table}[h]
    \centering
    \caption{Image distribution by class and dataset (training, validation, and test) at 108 Military Central Hospital (2018-2024)}
    \begin{tabular}{lccc}
        \toprule
        Class & Train & Validation & Test \\
        \midrule
        \gls{BENIGN} & 353 & 68 & 61 \\
        \gls{INDET_SUS} & 368 & 77 & 96 \\
        \gls{MALIGNANT} & 541 & 125 & 115 \\
        \bottomrule
    \end{tabular}
    \label{table1}
\end{table}

The distribution of images across these sets is detailed in table \ref{table1}. This stratification ensures that each subset reflects the original class distribution, maintaining statistical relevance across the training, validation, and testing phases.

In addition to the primary training and testing dataset, a prospective external validation set comprising 1,015 \gls{FNAB} images was collected from Hung Viet Hospital, the largest specialized thyroid cancer diagnosis and treatment center in Northern Vietnam, starting in July 2024 during the deployment phase. This external dataset serves as a critical benchmark to evaluate the model's performance in an independent clinical setting, providing essential evidence for external validity and generalizability. The prospective collection methodology closely mirrors real-world clinical use, where the model will encounter new, unseen data.
However, regarding the decision-making process at Hung Viet Hospital, it is crucial to emphasize that the cytopathologists' final diagnoses were intended to be entirely independent of the model's predictions. This was to prevent any potential influence of the model's output on the specialists' assessments, thereby ensuring the integrity of the 'ground truth' used for external validation. 

\subsection{Methodological}

During our research, we identified several key challenges:

\begin{itemize}
    \item \textbf{High Dimensionality:} While our dataset is relatively substantial within the context of current research in this domain, we encountered the dual challenge of high-dimensional input data coupled with moderate class representation (300-500 samples per class). This dimensionality-to-data ratio becomes particularly demanding when contrasted with classical computer vision tasks such as handwritten digit recognition (MNIST), where 60,000 samples per class compensate for lower-dimensional $28 \times 28$ grayscale inputs. Attempting to preserve the original image size to ensure all information is processed by the network exacerbates this issue.
    \item \textbf{Computational Burden (Full-Size Images):} Maintaining the original image dimensions leads to the development of large and inefficient models due to the necessity of wide and deep initial layers. Conversely, attempts to reduce input dimensionality through resizing or cropping inherently result in information loss.
    \item \textbf{Information Loss (Resizing):} Downscaling images to accommodate network input requirements can blur critical diagnostic features essential for accurate classification.
    \item \textbf{Contextual Information Loss (Cropping):} Extracting localized image regions to preserve fine details sacrifices the broader contextual information crucial for comprehensive diagnosis.
\end{itemize}

\begin{figure}
    \centering
    \includegraphics[width=1\linewidth]{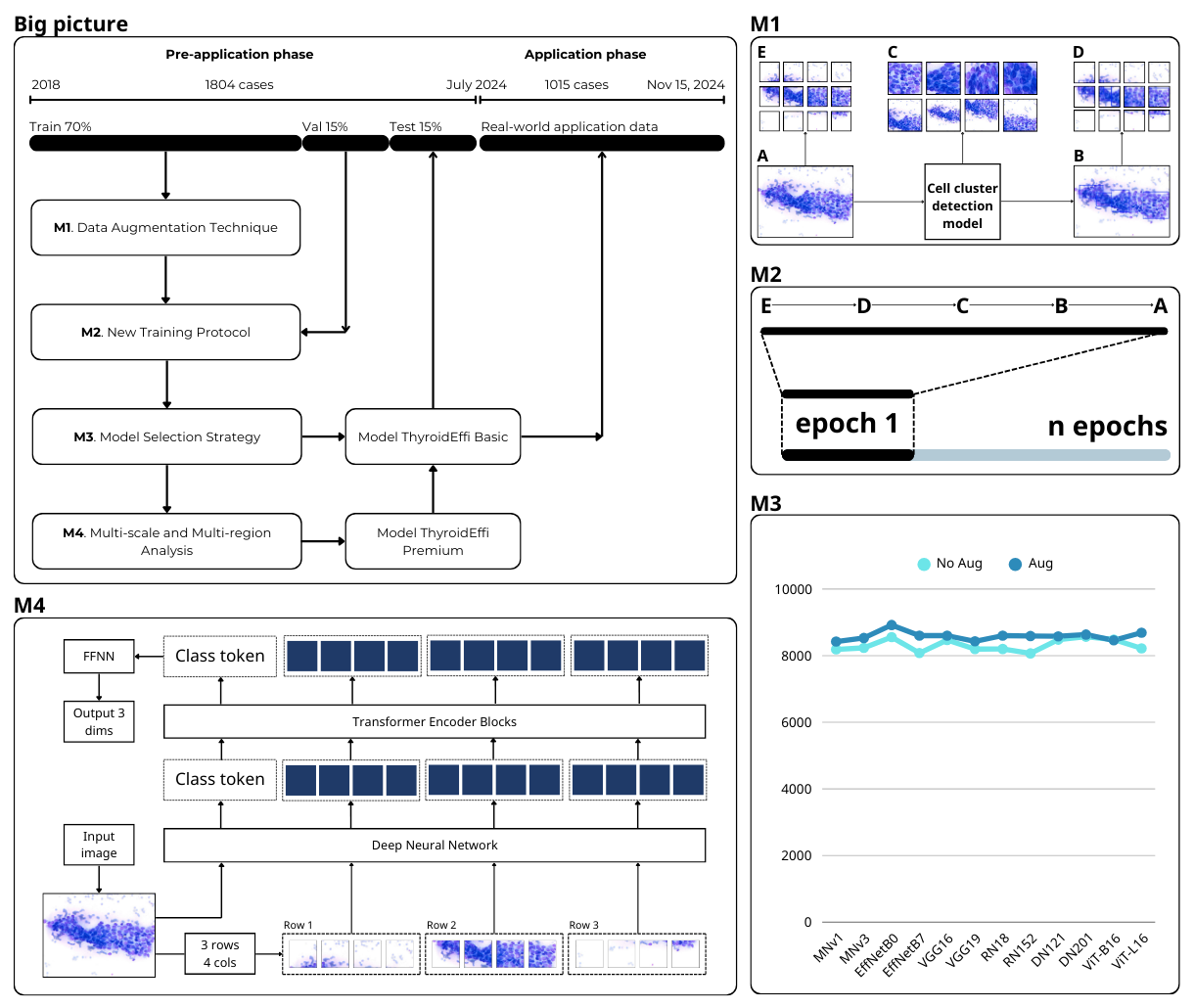}
    \caption{Visual overview of our methodology, encompassing data pre-processing, model training with novel techniques (M1-M4). To avoid any misunderstanding, it is crucial to note that the techniques described in M1 were exclusively applied to the training dataset, consequently meaning that M2's intended function is solely relevant during the training phase. The internal validation, test, and external validation datasets were strictly reserved for their designated purposes: preventing model overfitting (internal validation) and evaluating the trained model's performance (test and external validation). Following model development and testing before July 2024 at 108 Military Central Hospital, the model entered a deployment phase with extended testing on an independent dataset at Hung Viet Hospital.}
    \label{fig3}
\end{figure}

To address these challenges while achieving our stated objectives, we designed a novel four-module pipeline (M1 to M4), detailed visually in Figure \ref{fig3}. These four seamlessly integrated modules work synergistically to mitigate the aforementioned limitations in the following ways:

\begin{itemize}
    \item \ref{method_m1}{ M1}: Employs a sophisticated data augmentation strategy on the training set. This strategy generates diverse data variations, encompassing the preservation of global image characteristics (image sets A, B) and a targeted focus on critical local features (image sets D, E, and particularly image set C). Set C specifically emphasizes high cell density regions containing crucial morphological information for diagnostic decisions, while simultaneously reducing noise to facilitate faster and more effective model learning.
    \item \ref{method_m2}{ M2}: Enables the model to effectively learn and retain both global and local image characteristics. This is achieved through a sequential learning process that progresses from localized, cropped image regions (lacking contextual information) to downsized full-image views (preserving global context but potentially losing fine-grained local details in some areas).
    \item \ref{method_m4}{ M4}: Facilitates the integration of both global and local feature information to enable more accurate diagnostic decisions. The core idea is to leverage the understanding of global characteristics to contextualize the information derived from local features, leading to a more informed and comprehensive decision-making process through multi-region, multi-scale analysis.
    \item \ref{method_m3}{ M3}: Leverages the synergistic combination described in M1, M2, and M4, allowing the model to effectively learn and utilize both local and global features without the necessity of maintaining the original input image size. This significantly reduces model size and deployment costs. Furthermore, it simplifies the inference process by eliminating the need for manual region-of-interest extraction or pre-defining grid regions for classification, as well as bypassing the complex and computationally intensive ensemble techniques (like \gls{SVM} or AdaBoost) previously employed. Module M3 focuses on selecting the most lightweight \gls{CNN} architecture that maintains high classification performance.
\end{itemize}

Detailed information regarding the techniques employed in \ref{method_m1}{ M1}, \ref{method_m2}{ M2}, \ref{method_m3}{ M3}, and \ref{method_m4}{ M4} is provided, with further comprehensive details available in Appendices \ref{appendixD} and \ref{appendixE}.

\subsubsection{Data Augmentation Technique (M1)}
\label{method_m1}

The primary goals of our data augmentation technique are:
\begin{itemize}
    \item Addressing the issue of data scarcity and ensuring diverse sample representation in the dataset. The limited size of medical image datasets is a common challenge, potentially hindering model generalization.
    \item Proposing a novel data augmentation method that enhances the model’s ability to learn key features while minimizing the influence of noisy or irrelevant data. Augmentation should not simply increase the quantity of data but also improve its quality for learning.
\end{itemize}

Our approach employs sophisticated data augmentation tailored to improve model robustness across varying classes by focusing on key areas within the original images. This strategy aims to emulate the variability encountered in real-world clinical settings, such as variations in staining, illumination, and cellular morphology. As shown in Figure \ref{fig3}, M1, our data augmentation strategy increases the training dataset size by a factor of 34, using five sets:
\begin{itemize}
    \item Set A ($\times 1$): The original image with dimensions 1024x768 is processed using the YOLOv10 model \cite{wang2024yolov10} to detect cell clusters.
    \item Set B ($\times 1$): One image is generated by overlaying the top eight bounding boxes with the highest probabilities on the original image. This concentrates the model's attention on regions identified as likely containing diagnostically relevant cellular clusters.
    \item Set C ($\times 8$): Comprises eight regions cropped from the bounding boxes with the highest probabilities of containing densely clustered cells. If fewer than eight bounding boxes are detected, the remaining regions are randomly selected from a supplementary set of predefined cropped regions. These supplementary regions consist of six images (512x512 pixels) and two larger images (768x768 pixels), ensuring a balance between capturing fine details and broader contexts.
    \item Sets D ($\times 12$) and E($\times 12$): Comprise 12 images each, cropped from set B and the original images in set A using a uniform grid size of 256x256 pixels. This ensures uniformity in size, eliminates overlaps, and optimizes coverage of the entire image, thereby enhancing diversity in the training dataset. This strategy forces the model to learn from both focused cellular regions and broader contextual information, improving its ability to generalize to diverse image presentations.
\end{itemize}

For cell cluster detection, 120 images (in the training set) were manually annotated to label concentrated cell clusters. These labeled images were utilized to train the YOLOv10 model, which serves exclusively for the training data augmentation process. It is not employed during testing or inference, thereby streamlining the workflow and avoiding computational overhead during diagnosis. The YOLOv10 model was selected for its balance of speed and accuracy in object detection tasks, making it suitable for augmenting the training data without introducing significant computational bottlenecks. To address potential concerns about data leakage due to manual curation, strict protocols were followed during annotation, ensuring that the annotators were blinded to the final diagnostic labels of the images. The annotation process focused solely on identifying cell clusters based on morphological characteristics, without any reference to the Bethesda category.

\subsubsection{New Training Protocol (M2)}
\label{method_m2}

This training protocol builds upon the principle that patterns learned last are better retained and more effective in neural networks or human cognition. Thus, placing patient-level samples (Set A) at the end of each epoch enhances learning efficacy. By strategically prioritizing the learning of full-image context towards the end of each epoch, the model is encouraged to consolidate its understanding of the overall diagnostic landscape, ultimately improving its ability to make accurate patient-level classifications.

During the training process, each individual augmented image or cropped region generated through the M1 strategy (from sets A, B, C, D, and E) was treated as a separate training sample. These samples were assigned the overall diagnostic label (\gls{BENIGN}, \gls{INDET_SUS}, or \gls{MALIGNANT}) of their original full image.
A notable feature of our training approach is the structured sequence of images within each epoch, adhering to a defined data augmentation hierarchy. Images are processed sequentially from sets E, D, C, B, to A, as shown in figure \ref{fig3}, M2. This arrangement enables the model to progressively learn features at multiple scales, beginning with high-noise images (set E) and advancing to low-noise sets. The specific order of images in the training sequence (E to A) is designed to gradually increase the complexity and contextual richness of the training data. By starting with smaller, high-noise regions, the model is encouraged to focus on identifying fundamental cellular features. As the training progresses through sets D, C, and B, the model is exposed to increasingly larger and more complete image regions, allowing it to integrate these local features into a broader understanding of the overall cytological context.

Concluding each sequence with set A refines the model’s predictive capabilities on real-world-like images, eliminating the need for resource-intensive preprocessing while ensuring reliable outputs. Furthermore, this protocol enhances multi-scale comprehension by integrating insights from localized features in smaller regions and broader patterns from larger areas. This contributes to improved diagnostic accuracy and reliability at the patient level. This curriculum learning-inspired approach improves the model's ability to handle the inherent variability in cytological samples, ultimately contributing to more accurate and reliable diagnostic predictions.

While individual cropped regions from sets C, D, and E may not encompass all diagnostic features of the full image, assigning the full image's label to these localized views is a deliberate strategy. This approach, combined with the curriculum learning sequence of M2 (progressing from localized crops to full images), trains the model to identify crucial cellular and tissue patterns even in limited views, while the subsequent training on broader contexts (sets B and A) allows the model to integrate these local findings for accurate patient-level classification, effectively mitigating the risk of misinterpretation based solely on partial information. Especially when combined with \href{method_m4}{M4}, this allows for the integration of both small local image features (which are often lost if images are resized to smaller dimensions to reduce deep learning model size) and global, large-scale features (which are often lost if only selected regions are cropped from the image). Ultimately, learning both local and global level features in \href{method_m2}{M2} and combining them in \href{method_m4}{M4} leads to better prediction quality by utilizing more feature information.

\subsubsection{Model Selection Strategy (M3)}
\label{method_m3}

The goal of this model selection strategy is to optimize both classification performance and computational efficiency, addressing the complexities of multi-class classification in the context of imbalanced datasets. Our approach prioritizes the inclusion of diverse feature representations in each epoch, ensuring that the model receives a well-rounded learning experience across all target classes.  This focus on both accuracy and efficiency reflects the need for a model that can be readily deployed in resource-constrained clinical settings.

In this study, we evaluated twelve models from six distinct architectural families to identify the optimal model for our objectives. The models were carefully assessed across multiple criteria, including classification accuracy (measured by F1-score), structural simplicity (measured by the number of parameters), and inference efficiency (measured by inference time), with the aim of striking a balance between computational cost and diagnostic reliability. The emphasis on F1-score reflects a desire to balance precision and recall, ensuring that the model is both accurate and sensitive to the presence of malignancy.

The selected models span various complexities, from highly parameterized deep networks to lightweight versions of the same architecture. This range allows for an in-depth evaluation of trade-offs in terms of computational efficiency, model complexity, and performance accuracy, particularly in handling the challenges of imbalanced datasets in multi-class classification. The models considered include traditional \gls{CNN} such as VGG \cite{simonyan2014vgg}, ResNet \cite{he2016resnet}, MobileNet \cite{howard2017mobilenets}, DenseNet \cite{huang2017densely}, EfficientNet \cite{tan2019efficientnet}, as well as Transformer-based architectures like Vision Transformer (ViT) \cite{dosovitskiy2020image}, offering a broad spectrum for comparison. Additionally, customized variants with adjusted loss functions to address class imbalance \ref{appendixE}, such as class weighting techniques, were explored. This adjustment is crucial to mitigate model bias toward majority classes, a common challenge in imbalanced datasets.

This evaluation is designed to provide critical insights into which model architectures and configurations deliver the best balance between computational efficiency and classification accuracy, particularly when dealing with complex, imbalanced data. The findings will help guide the selection of a model that not only meets performance benchmarks but also aligns with resource constraints common in clinical and real-time diagnostic settings.

\subsubsection{Multi-scale and Multi-region Analysis (M4)}
\label{method_m4}

This analysis aims to refine diagnostic accuracy by utilizing a single-model approach that assimilates information from multiple sub-regions of the original image. This method is inspired by Transformer-based architectures \cite{vaswani2017attention}, where sub-regions of an image are treated as individual tokens, enhancing the model’s ability to capture detailed features across the entire image. This approach allows the model to capture both local and global contextual information, improving its ability to differentiate between the subtle morphological differences that characterize the different Bethesda categories.

Our method follows a structure similar to the Vision Transformer (ViT), but introduces three key distinctions. First, rather than relying on basic convolutional layers to map cropped image regions into vectors, we utilize a deep network trained across multiple scales and sub-regions of the image. Second, we increase the number of extracted sub-regions to 12, arranged in a 3x4 grid, as opposed to the typical 9 regions. Third, the class token, representing the overall classification of the image, is initialized using the output from the deep network applied to the original image, rather than a random value. This class token is then refined through successive iterations as it correlates with the 12 cropped patches. The refined token is passed through Transformer encoder layers and a \gls{FFNN} to generate the final three-class classification output.

To explore the benefits of automatic, grid-based region selection in thyroid biopsy image analysis, we implemented a Transformer-inspired model that interprets each image as a sequence of tokens. Each token corresponds to a unique region in a fixed grid layout, enabling multi-region inference in a single pass and simplifying the analysis process. The adoption of a grid-based approach ensures that all areas of the image are considered, avoiding potential bias introduced by manual \gls{ROI} selection.

Figure \ref{fig3}, M4 provides an overview of the model design, detailing the following key steps in Appendix \ref{appendixD8}.

\section{Results}

When comparing with previous research methods applied to our dataset, we note that studies in this area often lack direct comparison with one or more previous implementations. This stems from several key differences:

\begin{itemize}
    \item \textbf{Variations in Datasets and Preprocessing:} Some studies manually extract image fragments for individual fragment-level prediction. Replicating this on our dataset would be infeasible or prohibitively expensive, especially considering our aim for automated inference, where manual fragment extraction is impractical and costly for real-world deployment. Other research employs multi-stage analysis, such as training a network to segment the image into a grid and identify relevant grid cells for prediction, followed by another network to classify these selected regions. This approach is also cost-intensive, requiring extensive annotation of our entire dataset to determine significant grid cells. Furthermore, the multi-stage processing pipeline contradicts our research objective of minimizing specialized steps during inference.
    \item \textbf{Different Approaches to Patient-Level Classification Aggregation:} Prior studies aggregate classification results at the patient level from predictions on extracted fragments or selected grid regions. Some even ensemble results from multiple complex models to arrive at a final prediction, which increases inference time and computational cost in real-world deployment scenarios.
\end{itemize}

Therefore, to provide a meaningful comparison with previous approaches, we will refrain from manual image cropping, manual labeling of significant grid regions, and ensemble techniques. Instead, our comparison will focus on evaluating our implementation against the performance of the component neural networks utilized in prior research. It logically follows that if our component network achieves superior results, any subsequent aggregation or ensembling would likely yield even better outcomes. Detailed comparison results using these component networks (\gls{CNN} architectures commonly employed in previous studies) can be found in the analysis presented in Section \ref{result_1}{result on test set}, specifically in Table \ref{table2}.

\subsection{ThyroidEffi Basic Model Performance on Test Set}
\label{result_1}

\begin{table}[h]
\centering
\caption{F1 Score on Test Set for Different Models and Configurations}
\begin{tabular}{lcccc}
\toprule
Model & F1 score (no\_aug) & F1 score (aug) & Delta (aug - no\_aug) \\
\midrule
MNv1 & 0.8188 & 0.8423 & \cellcolor{green!25}0.0235 \\
MNv3 & 0.8234 & 0.8530 & \cellcolor{green!25}0.0296 \\
\midrule
EffNetB0 & 0.8555 & \textbf{0.8919} & \cellcolor{green!25}0.0364 \\
EffNetB7 & 0.8078 & 0.8600 & \cellcolor{green!25}\textbf{0.0522} \\
\midrule
VGG16 & 0.8468 & 0.8600 & \cellcolor{green!25}0.0132 \\
VGG19 & 0.8196 & 0.8430 & \cellcolor{green!25}0.0234 \\
\midrule
RN18 & 0.8200 & 0.8600 & \cellcolor{green!25}0.0400 \\
RN152 & 0.8067 & 0.8588 & \cellcolor{green!25}0.0521 \\
\midrule
DN121 & 0.8480 & 0.8580 & \cellcolor{green!25}0.0100 \\
DN201 & \textbf{0.8569} & 0.8633 & \cellcolor{green!25}0.0064 \\
\midrule
ViT-B16 & 0.8483 & 0.8460 & \cellcolor{red!25}-0.0023 \\
ViT-L16 & 0.8220 & 0.8690 & \cellcolor{green!25}0.0470 \\
\bottomrule
\end{tabular}
\label{table2}
\end{table}

Table \ref{table2} shows a comparison of macro F1 scores. We prioritized using the macro F1 score because it provides a balanced measure, ensuring accurate performance assessment across all three classification classes, and is less sensitive to data imbalance compared to overall accuracy, thus providing a more reliable representation of the model's overall effectiveness for all classes. The table compares the performance of common \gls{CNN} models replicated from previous studies (shown in column 2) with the performance of these models when implemented within our pipeline (shown in column 3). This comparison, excluding ViT (which is a new model architecture included due to its prevalence and performance in recent studies on large datasets), highlights the impact of our pipeline and training protocol on model performance. The $\times 34$ data augmentation approach led to substantial performance improvements across most models, except for ViT-B16 (highlighted in red). EfficientNetB0, trained with the augmented dataset, achieved the highest macro F1 score of 89.19\% and is designated as ThyroidEffi Basic.

\subsection{Comparison of ThyroidEffi Basic and ThyroidEffi Premium}

ThyroidEffi Basic achieved a macro F1 score of 89.19\%. By adopting a Transformer-based architecture and further refinement, ThyroidEffi Premium was developed, achieving a macro F1 score of 89.77\%. This represents an improvement of 0.58\% in macro F1 score. While the improvement is moderate, it demonstrates a trade-off between enhanced performance and increased computational cost associated with the more complex Transformer architecture.

\begin{table}[h]    
    \centering
    \caption{Confusion matrix of the ThyroidEffi Basic model on test set}
    \begin{tabular}{lrrrr}
    \toprule
    & \multicolumn{4}{c}{Predicted Label} \\
    \cmidrule(lr){2-5}
    True Label & \multicolumn{1}{c}{\gls{BENIGN}} & \multicolumn{1}{c}{\gls{INDET_SUS}} & \multicolumn{1}{c}{\gls{MALIGNANT}} & \multicolumn{1}{c}{Total} \\
    \midrule
    \gls{BENIGN} & \cellcolor{blue!25}57 & \cellcolor{blue!2}2 & \cellcolor{blue!2}2 & 61 \\
    \gls{INDET_SUS} & \cellcolor{blue!0}0 & \cellcolor{blue!85}85 & \cellcolor{blue!11}11 & 96 \\
    \gls{MALIGNANT} & \cellcolor{blue!2}2 & \cellcolor{blue!15}15 & \cellcolor{blue!98}98 & 115 \\
    \midrule
    Total & 59 & 102 & 111 & 272 \\
    \bottomrule
    \end{tabular}
    \label{table3}
\end{table}

The confusion matrix for the ThyroidEffi Basic model (EfficientNet B0) on the test set is presented in table \ref{table3}.

The ThyroidEffi Basic model also achieved \gls{AUC} scores of \gls{BENIGN}: 0.98, \gls{INDET_SUS}: 0.95, and \gls{MALIGNANT}: 0.96 on the test set.

\subsection{Real-World Application Results}

The ThyroidEffi Basic model was deployed at Hung Viet Hospital, analyzing 1,015 patients. The performance metrics are summarized below:
\begin{itemize}
    \item \gls{AUC}: \gls{BENIGN}: 0.9495, \gls{INDET_SUS}: 0.7436, \gls{MALIGNANT}: 0.8396.
    \item The classification report and confusion matrix for the real-world application data are shown in table \ref{table4}  and table \ref{table5}, respectively.
\end{itemize}

\begin{table}[h]
    \centering
    \caption{Performance evaluation of the ThyroidEffi Basic model on real-world application data}
    \begin{tabular}{lcccc}
    \toprule
    & \gls{BENIGN} & \gls{INDET_SUS} & \gls{MALIGNANT} & Macro Avg \\
    \midrule
    F1-Score & 0.84 & 0.49 & 0.70 & 0.68 \\
    \bottomrule
    \end{tabular}
    \label{table4}
\end{table}

\begin{table}[h]
    \centering
    \caption{Confusion matrix for the ThyroidEffi Basic model on real-world application data}
    \begin{tabular}{lrrrr}
    \toprule
    & \multicolumn{3}{c}{Predicted Label} & \\
    \cmidrule(lr){2-4}
    True Label & \multicolumn{1}{c}{\gls{BENIGN}} & \multicolumn{1}{c}{\gls{INDET_SUS}} & \multicolumn{1}{c}{\gls{MALIGNANT}} & \multicolumn{1}{c}{Total} \\
    \midrule
    \gls{BENIGN} & \cellcolor{blue!85}261 & \cellcolor{blue!15}30 & \cellcolor{blue!5}9 & 300 \\
    \gls{INDET_SUS} & \cellcolor{blue!20}44 & \cellcolor{blue!50}143 & \cellcolor{blue!40}128 & 315 \\
    \gls{MALIGNANT} & \cellcolor{blue!10}16 & \cellcolor{blue!35}92 & \cellcolor{blue!75}292 & 400 \\
    \midrule
    Total & 321 & 265 & 429 & 1015 \\
    \bottomrule
    \end{tabular}
    \label{table5}
\end{table}

\subsection{Real-World Deployment}

Our application is currently being deployed at partner institutions via \href{https://harito.id.vn/solution}{https://harito.id.vn/solution}. General statistics indicate low deployment costs due to the system's lightweight nature. For instance, a CPU with specifications similar to a 12th Gen Intel i5-12450H (12 cores @ 4.400GHz) can process an average of 1000 cases in just 30 seconds.

\section{Discussion}
\label{discussion}

\subsection{Cost-Effectiveness and High Performance of the System}

As demonstrated, the system features a lightweight deployment (only 4 million parameters for the Basic model and 4.5 million parameters for the Premium model) with minimal costs while maximizing the classification performance across all three categories. This synergy enhances the practical deployability of the system in real-world settings.

It is important to clarify that during inference and real-world deployment, the model receives only the raw or minimally pre-processed image data as input. It does not have access to or receive any pre-existing information about the pathologist's ground truth diagnosis or the original Bethesda category (I-VI) of the image. The model's output is solely its prediction of the likelihood distribution across the three defined clinical management groups based on the learned features from the image.

\subsection{Impact of Data Augmentation and Model Complexity}

The results in table \ref{table2} demonstrate the significant impact of data augmentation on model performance. The augmentation strategy improved F1 scores across most models, indicating that increased data diversity helps models learn more robust features and generalize better. However, ViT-B16 showed a decrease in performance with augmentation, likely due to its lower capacity compared to ViT-L16. ViT-B16 might be more susceptible to noise introduced by the augmented data, struggling to distinguish relevant features. Conversely, models with moderate complexity, such as EfficientNet, ResNet, and VGG, benefited most from augmentation, achieving a better balance between learning from diverse data and avoiding overfitting. This suggests that model capacity plays a crucial role in leveraging the benefits of data augmentation.

The observation that models with excessively low (MobileNet) or high (DenseNet, ViT) parameter counts do not always achieve the best performance further supports the importance of balanced complexity. EfficientNet, VGG, and ResNet, with their moderate parameter counts, demonstrated superior performance, suggesting an optimal balance between model capacity and generalization ability for this specific task.

\subsection{Model Explainability and Visualization}

\begin{figure}
    \centering
    \includegraphics[width=1\linewidth]{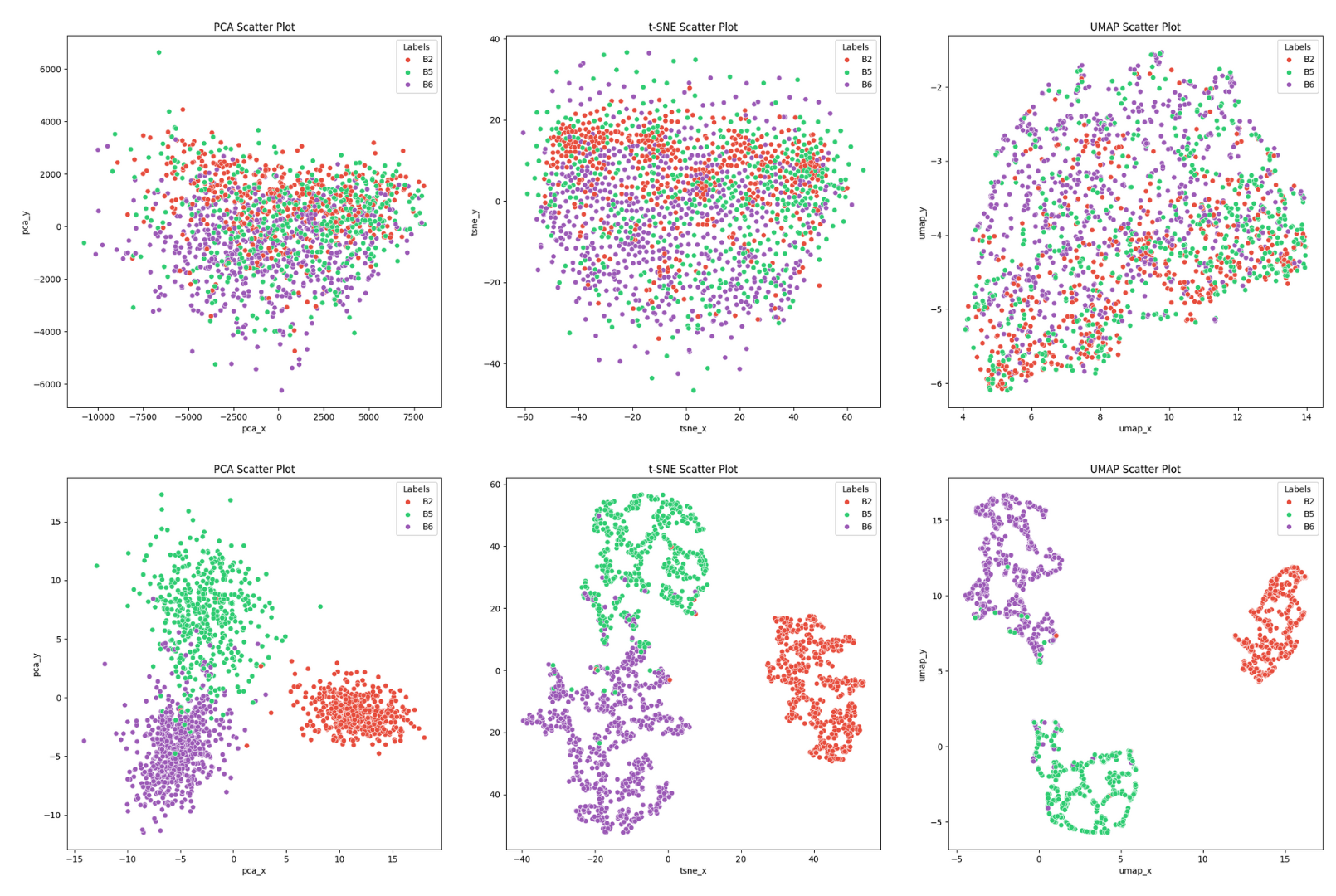}
    \caption{Visualization of data distribution before and after applying the ThyroidEffi Basic model for classifying \gls{BENIGN}, \gls{INDET_SUS}, and \gls{MALIGNANT} categories: \\
- Top row: Original dataset reduced to 2D using Principal Component Analysis (\gls{PCA}) \cite{dr_PCA}, t-distributed Stochastic Neighbor Embedding (\gls{t-SNE}), and Uniform Manifold Approximation and Projection (\gls{UMAP}) \cite{dr_tSNE_UMAP}, showing overlapping clusters and poor separability among classes. \\
- Bottom row: Dataset transformed by ThyroidEffi Basic into a 3-dimensional latent space (representing probabilities for \gls{BENIGN}, \gls{INDET_SUS}, \gls{MALIGNANT}) and reduced to 2D. Clear cluster separation highlights the model’s ability to extract discriminative features.}
    \label{fig4}
\end{figure}

Figure \ref{fig4} visually demonstrates the effectiveness of the ThyroidEffi Basic model in transforming complex image features into a more discriminative latent space. The clear separation of clusters in the transformed space highlights the model's ability to extract relevant features for classifying \gls{BENIGN}, \gls{INDET_SUS}, and \gls{MALIGNANT} categories. This visualization provides insights into the model's decision-making process and supports its validity.

\begin{figure}
    \centering
    \includegraphics[width=1\linewidth]{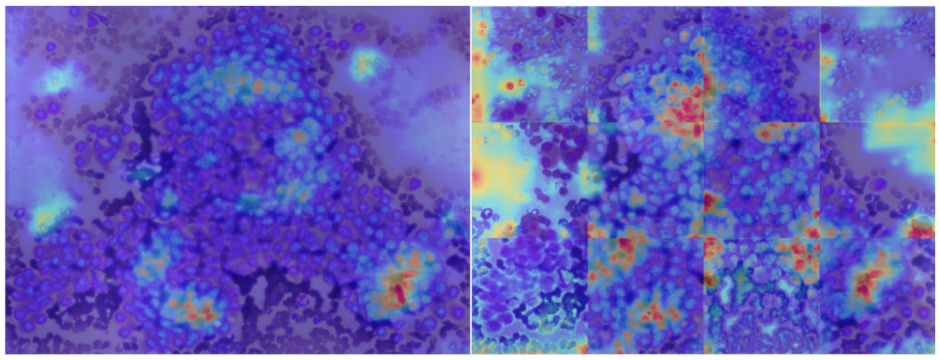}
    \caption{\gls{Grad-CAM} of a slide image (left) and 12 sub-images (right) after pass ThyroidEffi Basic}
    \label{fig5}
\end{figure}

The use of \gls{Grad-CAM} \cite{selvaraCAMju2017gradCAM} (figure \ref{fig5}) further enhances model explainability by highlighting the image regions that contribute most to the model's predictions. This allows clinicians to understand why the model makes certain classifications, increasing trust and facilitating validation of the automated decisions.

\subsection{Real-World Application Challenges}
\label{discussion_real_world}

Real-world deployment and external validation observations revealed a performance disparity compared to initial testing. Specifically, the model demonstrated a tendency to misclassify \gls{INDET_SUS} cases, sometimes assigning them to the \gls{MALIGNANT} category, which contributed to a lower overall performance on the external validation set. While both the training/initial testing dataset from 108 Military Central Hospital and the external validation dataset from Hung Viet Hospital represent real-world data, variations in performance between institutions are a known challenge in medical imaging due to differences in data distribution.

Potential causes for these observed performance differences and the model's generalizability challenges likely include:
\begin{itemize}
    \item \textbf{Patient Demographics:} Differences in patient populations, prevalence of specific thyroid pathologies, or demographic factors between the two hospitals.
    \item \textbf{Imaging Protocols and Equipment:} Variations in ultrasound equipment models, imaging settings, and \gls{FNAB} acquisition protocols between the facilities can lead to distinct visual characteristics in the slide images.
    \item \textbf{Slide Preparation and Staining Variability:} Differences in laboratory procedures for sample processing, slide preparation, and staining techniques can introduce variations in cellular morphology, background artifacts, and staining intensity.
    \item \textbf{Pathologist Variability:} Despite standardized guidelines like \gls{TBSRTC}, subtle differences in subjective interpretation and diagnostic thresholds among pathologists at different institutions can influence ground truth labeling. This is particularly pronounced for indeterminate categories, which are inherently more subjective within \gls{TBSRTC} and prone to higher inter-observer variability, thus contributing to less consistent ground truth across institutions.
    \item \textbf{Temporal Data Drift:} Changes in equipment calibration, protocols, or even disease characteristics over time could lead to the external validation data having subtly different feature distributions than the training data.
\end{itemize}
The lower performance observed on the external validation set indicates that the current model's generalizability across different clinical sites and potentially over time presents a challenge that warrants further attention.

Future research will focus on further elucidating the specific contributions of these potential causes to model performance variability. We also plan to continuously improve the model's robustness and generalizability through strategies such as continuous learning, which can help the model adapt to and capture changes in data distribution over time and across different clinical environments.

\section{Conclusions}

\subsection{Feature research}

This study demonstrates the effectiveness of deep learning models, particularly the ThyroidEffi Basic model, for automated analysis of thyroid cytology biopsy images. Despite having only \textbf{4 million parameters}, ThyroidEffi Basic achieved high accuracy on the test set and delivered promising results in a real-world clinical deployment, highlighting its efficiency in both performance and computational cost. The model’s lightweight architecture enables faster inference and lower hardware requirements, making it well-suited for real-time clinical applications. The use of data augmentation and careful selection of model complexity were crucial for achieving optimal performance without unnecessary computational overhead. Additionally, the integration of visualization techniques and \gls{Grad-CAM} further enhanced model explainability and facilitated clinical validation, ensuring both reliability and interpretability in medical decision-making.

\subsection{Limitations and Future Work}

While the model exhibited a tendency to misclassify some Bethesda V (Suspicious for Malignancy) cases as Bethesda VI (Malignant) in the real-world setting, this observation reflects the inherent diagnostic challenges in distinguishing between these categories and underscores the necessity for continuous refinement. Future work will prioritize addressing this limitation by integrating more diverse real-world data, with findings planned for publication in subsequent ThyroidEffi 1.x versions. Furthermore, we intend to enhance the model's explainability by developing methodologies to highlight specific regions of interest on cytology images, identifying key cytological features such as abnormal cell morphology, sharp nuclear angles, and clustered cellular arrangements that contribute to the model's diagnostic decisions. These advancements are slated for release in ThyroidEffi 2.x versions. This improved interpretability will provide clinicians with a clearer rationale behind the model's predictions, fostering greater trust and facilitating more precise diagnostic assessments.

Future iterations, under the ThyroidEffi 3.x roadmap, will focus on improving diagnostic quality in cases of suboptimal image clarity, as well as expanding the model's diagnostic capabilities to other cytopathology domains (e.g., breast cancer), leveraging the transferable expertise gained from thyroid cytopathology.

The developed end-to-end system holds significant potential to improve the efficiency and accuracy of thyroid cancer diagnosis, ultimately enhancing clinical decision-making and patient outcomes.

\section{Summary table}

What was already known on the topic:
\begin{itemize}
    \item \textbf{Binary Classification Limitation}: Many studies focus on binary classification (benign/malignant), neglecting the full multi-class categorization of \gls{TBSRTC}.
    \item \textbf{Dataset Size Constraints}: Limited dataset sizes hinder generalizability and model robustness.
    \item \textbf{Lack of Patient-Level Predictions}: Most models predict at the fragment/image level, lacking integrated patient-level insights.
    \item \textbf{Computational Complexity}: High computational complexity makes real-time clinical deployment challenging.
\end{itemize}

What this study added to our knowledge:
\begin{itemize}
    \item \textbf{Data Augmentation Technique}: Addresses the dataset size constraint by generating synthetic data, enhancing model robustness and generalizability. This technique focuses on augmenting key regions within the original images, minimizing the impact of noise.
    \item \textbf{New Training Protocol}: Enables effective patient-level image classification and mitigates label imbalance. By strategically prioritizing certain records, especially patient-level images, during training, the model learns effectively from all classes and improves patient-level diagnostic accuracy.
    \item \textbf{Model Selection Strategy}: Optimizes the balance between classification performance and computational efficiency. A comprehensive evaluation of twelve models from six architectural groups was performed, including customized variants with modified loss functions to address class imbalance.
    \item \textbf{Multi-scale and Multi-region Analysis}: Improves diagnostic accuracy by aggregating information from multiple image regions, mimicking a Transformer-based architecture. This approach captures detailed features across the entire sample within a single inference flow, addressing the lack of patient-level predictions.
\end{itemize}

\section*{Availability of data and materials}

The source code for the proposed method is publicly available on GitHub at \url{https://github.com/Harito97/Research_ThyroidFNA_ClassAI}.

Due to patient privacy restrictions, the datasets used in this study cannot be made publicly available. However, they can be obtained from the corresponding author upon reasonable request and after signing a data sharing agreement with the participating institutions.

Furthermore, this research has been implemented as a cytopathology analysis tool, which is accessible at
\url{https://harito.id.vn/solution}.

\section*{Ethics statement}

This study was conducted in accordance with the ethical guidelines of the Institutional Review Board at 108 Military Central Hospital, Hanoi, Vietnam, the Research Ethics Committee at Hung Viet Hospital, Hanoi, Vietnam, and the Research Ethics Committee of Vietnam National University, Hanoi, Vietnam, with written informed consent obtained from all participants. All participants provided written informed consent in accordance with the Declaration of Helsinki. The study protocol was reviewed and approved by the relevant ethics committees at 108 Military Central Hospital, Hung Viet Hospital, and Vietnam National University. Participants provided their written informed consent to participate in this research.

\section*{CRediT authorship contribution statement}

\textbf{Hai Pham-Ngoc}: Conceptualization, Methodology, Visualization, Software, Writing - Original Draft. \textbf{De Nguyen-Van}: Formal analysis, Investigation, Data Curation, Writing  - Review \& Editing. \textbf{Dung Vu-Tien}: Validation, Supervision, Writing - Review \& Editing. \textbf{Phuong Le-Hong}: Resources, Supervision, Project Administration, Writing - Review \& Editing.

\section*{Funding}

This research did not receive any specific grant from funding agencies in the public, commercial, or not-for-profit sectors.

\section*{Declaration of competing interest}

The authors declare that they have no known competing financial interests or personal relationships that could have appeared to influence the work reported in this paper.

\printglossaries

\bibliographystyle{unsrtnat}
\bibliography{main.bib}

\newpage

\appendix

\section*{Appendix}

The criteria for evaluating the quality of a research study (particularly those employing AI in medical image diagnosis) are listed sequentially by session below.

\section*{Appendix A. Problem Understanding}
\label{appendixA}

\subsection*{A.1 Study Population and Inclusion/Exclusion Criteria}

\textbf{Inclusion Criteria:}
\begin{itemize}
    \item Patients aged 18 years or older.
    \item Patients presenting with thyroid nodules warranting \gls{FNAB}, as determined by clinical evaluation and ultrasound findings.
    \item Availability of sufficient cytological material in \gls{FNAB} samples to allow classification according to the Bethesda System (categorized as \gls{BENIGN}, \gls{INDET_SUS}, or \gls{MALIGNANT}).
    \item Patients with no prior history of thyroid surgery or interventional procedures that might alter the cytological presentation.
\end{itemize}

\textbf{Exclusion Criteria:}
\begin{itemize}
    \item Patients under the age of 18.
    \item \gls{FNAB} samples that were deemed non-diagnostic due to inadequate cellularity or poor staining quality.
    \item Patients with a history of thyroid surgery or other therapeutic interventions that could compromise the interpretability of the \gls{FNAB}.
    \item Cases with ambiguous cytological findings that did not clearly fall within the predefined Bethesda categories.
\end{itemize}

\subsection*{A.2 Study Design}

The content has been secured in the main sections of the manuscript.

\subsection*{A.3 Study Setting}

The content has been secured in the main sections of the manuscript.

\subsection*{A.4 Data Source}

The content has been secured in the main sections of the manuscript.

\subsection*{A.5 Medical Task Description}

The content has been secured in the main sections of the manuscript.

\subsection*{A.6 Data Collection Process and Quality Considerations}

The collection process adhered to standardized clinical protocols for thyroid nodule evaluation, ensuring consistency and reliability. Key aspects of the data collection process include:

\textbf{Sampling and Slide Preparation:}
\begin{itemize}
    \item \gls{FNAB} procedures were performed by experienced clinicians following standard clinical guidelines.
    \item Each \gls{FNAB} sample was prepared on a slide and stained using a standardized Diff-Quick protocol to maintain consistency in cytological appearance.
\end{itemize}

\textbf{Imaging Protocol:}
\begin{itemize}
    \item Images were captured using an Olympus BX43 optical microscope equipped with a high-resolution digital camera.
    \item To ensure consistent image quality across all samples, imaging settings were standardized, including magnification (x40), lighting, exposure, and zoom.
\end{itemize}

\textbf{Quality Control Measures:}
\begin{itemize}
    \item Rigorous quality control procedures were implemented during data acquisition. All images were examined for key parameters including resolution, brightness, and contrast.
    \item Images that did not meet predefined quality thresholds were excluded from the dataset.
\end{itemize}

\textbf{Data Recording and Annotation:}
\begin{itemize}
    \item Each \gls{FNAB} image was associated with a unique patient identifier and labeled according to \gls{TBSRTC} (\gls{BENIGN}, \gls{INDET_SUS}, or \gls{MALIGNANT}) by expert cytopathologists.
    \item The data collection process was integrated into the hospital’s routine diagnostic workflow, ensuring that the samples were collected independently of any specific diagnostic hypothesis.
\end{itemize}

A similar standardized data collection and quality assurance process was followed for the external validation dataset at Hung Viet Hospital. This approach ensures that the datasets are robust, complete, and comparable across different clinical settings.

\section*{Appendix B. Data Understanding}
\label{appendixB}

\subsection*{B.1 Subject Demographics}

Demographic data were retrospectively extracted from the corresponding electronic medical records. The key demographic characteristics for the primary dataset (n = 1,804) are summarized below:

\begin{itemize}
    \item \textbf{Age:}  
    \begin{itemize}
        \item \textbf{Mean Age:} 47.5 years  
        \item \textbf{Standard Deviation (SD):} 12.3 years  
        \item \textbf{Median Age:} 46 years  
        \item \textbf{Inter-Quartile Range (IQR):} 38--56 years  
    \end{itemize}
    \item \textbf{Gender Breakdown:} Approximately 68\% of the patients were female and 32\% were male.
    \item \textbf{Main Comorbidities:}  
    \begin{itemize}
        \item Hypertension (22\%)  
        \item Diabetes Mellitus (15\%)  
        \item History of thyroid disorders (18\%)
    \end{itemize}
    \item \textbf{Ethnic Group:} 100\% of the subjects were of Vietnamese descent.
    \item \textbf{Socioeconomic Status:} No statistics available.
\end{itemize}

\subsection*{B.2 Gold Standard: Ground Truthing Process}

\gls{FNAB} samples were classified according to \gls{TBSRTC} guidelines by 3 board-certified cytopathologists with an average of 8 years of experience. Each sample was independently reviewed by at least two cytopathologists. 
Discrepancies in classification were resolved through discussion and consensus. 
In cases where consensus could not be reached, a third expert was consulted.
Image quality was controlled by ensuring all images met pre-defined criteria for resolution, brightness, and contrast. Any images failing to meet these criteria were excluded from the study.

\section*{Appendix C. Data Preparation}
\label{appendixC}

\subsection*{C.1 Outlier Detection and Analysis}

We did not perform explicit outlier detection in this study. Given the large size of our dataset and the use of robust data augmentation techniques, we anticipated that the model would be robust to a small number of potential outliers. Furthermore, the manual review of a subset of the data by expert cytopathologists during the ground truthing process helped to identify and correct any obvious mislabeled or corrupted images. Future work may explore more formal outlier detection methods.

\subsection*{C.2 Missing Data Management}

In this study, we proactively addressed the issue of missing data by manually removing incomplete or blurry images during the data acquisition process. Specifically:

\begin{itemize}
    \item \textbf{Incomplete Images:} During the image capture process, some images may not have fully recorded the slide due to technical errors or photographer oversight. We meticulously examined each image and excluded those that were incomplete, estimating this to be approximately 1\% of the total images.
    \item \textbf{Blurry Images:} Image quality can be affected by insufficient sharpness due to factors such as lighting conditions or camera settings. We carefully reviewed each image and excluded those that were blurry, estimating this to be approximately 3\% of the total images.
\end{itemize}

This manual removal was performed by the photographers themselves during the data collection process, ensuring the quality of the dataset before it was even considered for analysis. While we did not employ more sophisticated missing data handling methods, we believe that this manual removal helped to minimize the negative impact of missing data on the study results.

Note that images corresponding to Bethesda I samples were included in the \gls{INDET_SUS} category if they met the predefined quality criteria for cellularity and staining, while those deemed non-diagnostic due to inadequate material or poor quality were excluded as part of standard data preparation."

In the future, we may consider using more advanced missing data handling methods, such as:

\begin{itemize}
    \item \textbf{Automated Identification and Removal:} Developing algorithms to automatically identify and remove incomplete or low-quality image regions.
    \item \textbf{Data Imputation:} Replacing missing image regions with appropriate data, such as using interpolation algorithms or Generative Adversarial Networks (GANs).
\end{itemize}

\subsection*{C.3 Feature Pre-processing}

The images underwent two resizing steps before being input to the deep learning model.

\begin{enumerate}
    \item \textbf{Initial Resizing:} All images were initially resized to a consistent dimension of 1024x768 pixels. This standardization ensured uniformity in the input size \textbf{before} the data augmentation process.
    \item \textbf{Model Input Resizing:} Within the model architecture, the 1024x768 images were further resized to 224x224 pixels. This second resizing was performed to match the input requirements of the EfficientNetB0 model.
\end{enumerate}

No other pre-processing steps were performed. The pixel data, after these two resizing steps, served as the input features (224x224x3) for the convolutional neural network (EfficientNetB0).

\subsection*{C.4 Data Imbalance Analysis and Adjustment}

The dataset exhibited a class imbalance in the distribution of Bethesda categories.  The number of images per category was as follows:

\begin{itemize}
    \item \gls{BENIGN}: 482 images (26.72\%)
    \item \gls{INDET_SUS}: 541 images (29.99\%)
    \item \gls{MALIGNANT}: 781 images (43.29\%)
\end{itemize}

As can be seen, \gls{MALIGNANT} category had considerably more images than the other two categories. This imbalance could potentially bias the model towards the majority class.

To address this imbalance, we employed a weighted cross-entropy loss function during training. The weights for each class were calculated inversely proportional to their frequency in the dataset. This approach ensured that the model gave equal importance to all classes during training, mitigating the potential bias caused by the imbalanced distribution. 

\section*{Appendix D. Detailed Description of Models and Methods}
\label{appendixD}

\subsection*{D.1 Model Task}

The content has been secured in the main sections of the manuscript.

\subsection*{D.2 Model Output}

The model outputs a three-dimensional vector of probabilities, representing the predicted likelihood for each of the three Bethesda categories: \gls{BENIGN}, \gls{INDET_SUS}, and \gls{MALIGNANT}. Specifically, for a given input image, the model produces the following output:

\[
\mathbf{p} = [p_{\text{Benign}}, p_{\text{Indeterminate/Suspicious}}, p_{\text{Malignant}}]
\]

where:

\begin{itemize}
    \item $p_{Benign}$ represents the probability for the \gls{BENIGN} category.
    \item $p_{\text{Indeterminate/Suspicious}}$ represents the probability for the \gls{INDET_SUS} category.
    \item $p_{Malignant}$ represents the probability for the \gls{MALIGNANT} category.
\end{itemize}

These probabilities are normalized such that they sum to 1:

\[
p_{Benign} + p_{Indeterminate/Suspicious} + p_{Malignant} = 1
\]

The final classification decision is made by selecting the category with the highest predicted probability. That is, the image is assigned to the category $C$ such that:

\[
C = \arg\max_i p_i
\]

where $i \in \{Benign,  Indeterminate/Suspicious, Malignant\}$.

\subsection*{D.3 Data Splitting and Augmentation Protocol}
\label{appendixD3}

The details regarding data splitting have been thoroughly addressed in the main sections of the manuscript.

The data augmentation technique described in Section D.4 was exclusively applied to the training set. The validation and test sets were not augmented in any way and remained in their original form throughout the experiments. This ensures that the evaluation of the model's performance is conducted on unseen, unmodified data, providing a more accurate assessment of its real-world applicability.

An independent external validation set of 1,015 \gls{FNAB} images was collected from Hung Viet Hospital. This set was used exclusively for a final evaluation of the trained model's performance in a different clinical setting. It played no role in the training, validation, or hyperparameter tuning process.

\subsection*{D.4 Data Augmentation Technique: YOLOv10 Rationale and Overfitting Considerations}

Our augmentation strategy increases the training dataset size by a factor of 34 by extracting multiple sub-regions from each original image. YOLOv10 was chosen for cell cluster detection due to its well-documented attributes of being fast, lightweight, and easily fine-tuned for preprocessing tasks.

To mitigate the risk of overfitting due to the enlarged dataset-and to address potential class imbalance-the augmentation was applied uniformly across all classes. Additionally, regularization techniques were integrated into the training protocol, including:
\begin{itemize}
    \item \textbf{Dropout} and \textbf{weight decay} during model training,
    \item \textbf{Early stopping} (triggered if validation loss does not improve over 10 consecutive epochs), and
    \item A \textbf{weighted cross-entropy loss} that prioritizes minority classes.
\end{itemize}

\subsection*{D.5 New Training Protocol: Sequential Image Ordering (E $\rightarrow$ D $\rightarrow$ C $\rightarrow$ B $\rightarrow$ A)}

Our training protocol incorporates principles of curriculum learning by processing image subsets in a defined sequence: E, D, C, B, and A. In this sequence, sets E, D, and C consist of smaller cropped regions from the main image that capture localized, high-noise features, while sets B and A encompass larger portions or the entire image, providing global, patient-level context. This ordering leverages the observation that more recent training examples tend to have a stronger influence on model updates, allowing us to strategically prioritize the learning of increasingly complex and contextual features. By first challenging the model with localized, noisy regions and then progressively introducing broader, cleaner images, we aim to foster the effective learning and integration of both fine-grained details and comprehensive contextual features across multiple scales.

This sequential approach is firmly grounded in established curriculum learning concepts, where presenting examples in a structured manner can enhance learning efficiency and feature retention. The gradual transition from difficult local sub-regions to more informative global views is intended to enhance the retention of newly acquired features from simpler examples while building upon them with more complex ones, thereby improving multi-scale feature extraction and overall diagnostic performance by facilitating the model's ability to consolidate understanding across different levels of detail and context. This structured learning paradigm forms a core component of our novel training strategy.

\subsection*{D.6 Model Selection Strategy: Lightweight Model with High Performance}

Our evaluation encompassed 12 models across six architectural families, using multiple criteria such as macro F1 score, inference time, model size, and overall computational resource requirements. Notably, the EfficientNetB0 model emerged as the optimal candidate. When trained on the augmented dataset with the proposed E $\rightarrow$ A training order, EfficientNetB0 achieved the highest macro F1 score while maintaining a lightweight architecture with 4 million parameters. This balance of high performance and low computational cost renders it particularly suitable for real-time clinical applications.

\subsection*{D.7 EfficientNetB0 Architecture}
\label{appendixD4}

The ThyroidEffi Basic model utilizes the EfficientNetB0 architecture. EfficientNetB0 is a lightweight convolutional neural network known for its efficiency and strong performance on image classification tasks.  It employs a compound scaling method, uniformly scaling the network's depth, width, and resolution.  Key features of the architecture include:

\begin{itemize}
    \item \textbf{Mobile inverted bottleneck convolution (MBConv) blocks:} These blocks are the building blocks of EfficientNet and are designed for efficient computation on mobile devices. They utilize depthwise separable convolutions to reduce computational cost.
    \item \textbf{Squeeze-and-excitation (SE) blocks:}  These blocks learn channel-wise attention, allowing the network to focus on the most informative channels.
    \item \textbf{Compound scaling:} The network's depth, width, and resolution are scaled up together using a principled approach.  The scaling coefficients are determined by a neural architecture search.
\end{itemize}

The specific configuration of the EfficientNetB0 model used in this study was the standard configuration provided by the original authors.  
\subsection*{D.8 Multi-scale and Multi-region Analysis: Transformer-based Approach}
\label{appendixD8}
To capture both local and global features, we implemented a Transformer-inspired model for multi-scale and multi-region analysis. The approach processes each \gls{FNAB} image as follows:
\begin{enumerate}
    \item \textbf{Input Decomposition:} The original image \( I \) is partitioned into a 3$\times$4 grid, yielding 13 sub-images \( I_0, I_1, \dots, I_{12} \) that collectively cover the entire image.
    \item \textbf{Token Generation:} Each sub-image \( I_i \) is processed by a pre-trained classification model to generate a token \( T_i \) that encodes local feature information.
    \item \textbf{Dimensional Expansion:} Each token \( T_i \) is transformed into a higher-dimensional representation via a linear transformation, ensuring uniformity across tokens.
    \item \textbf{Transformer Encoder Processing:} The sequence of tokens is input into \( n \) Transformer encoder blocks (with a multi-head attention mechanism) to refine the feature representations.
    \item \textbf{Final Label Prediction:} The token corresponding to the primary region (typically \( T_0 \)) is passed through a \gls{FFNN} to produce the final three-dimensional output vector, representing the predicted class labels for the entire image.
\end{enumerate}

This grid-based, automated region selection method streamlines the inference pipeline by eliminating manual \gls{ROI} selection while ensuring that both localized and global diagnostic features are effectively captured.

\subsection*{D.9 Transformer-Inspired Module Architecture}
\label{appendixD9}

The ThyroidEffi Premium model incorporates a Transformer-inspired module for multi-scale and multi-region analysis.  This module builds upon the Vision Transformer (ViT) architecture but with the following key modifications:

\begin{itemize}
    \item \textbf{Sub-region Embedding:} Instead of using simple convolutional layers to embed the sub-regions, we use the pre-trained EfficientNetB0 model (from ThyroidEffi Basic) to generate embeddings for each of the 12 sub-regions.  This allows the module to leverage the features learned by the base model.
    \item \textbf{Number of Sub-regions:} We extract 12 sub-regions arranged in a 3x4 grid, rather than the 9 regions used in the standard ViT.
    \item \textbf{Class Token Initialization:}  The class token is initialized with the output of the EfficientNetB0 model applied to the original, full-sized image, as opposed to a random initialization.
    \item \textbf{Transformer Encoder Layers:} The embedded sub-region tokens and the initialized class token are then passed through 5 Transformer encoder layers.  Each encoder layer consists of multi-head self-attention and feed-forward neural networks.
    \item \textbf{Final Classification:}  The final classification is performed by passing the class token through a feed-forward neural network.
\end{itemize}

\section*{Appendix E. Validation}
\label{appendixE}

\subsection*{E.1 Data Splitting And Internal-External Model Validation Procedure}

\textbf{Internal Validation Procedure:}

We did not use k-fold cross-validation in this study for several reasons.  While k-fold cross-validation is a valuable technique, particularly for smaller datasets, it introduces additional computational cost.  With a dataset of 1804 images, we aimed for a balance between maximizing data usage and maintaining reasonable computational resources.  Furthermore, the benefits of k-fold cross-validation diminish with larger datasets, as a single, well-chosen train/validation/test split can provide a reliable estimate of performance. The robustness of our evaluation is further strengthened by the extensive external validation performed on an independent dataset.

\textbf{Data Splitting:}

As mentioned above, the data was split \textit{before} any normalization, standardization, or imputation. This is crucial to prevent data leakage. If these preprocessing steps were performed before splitting, information from the test set could inadvertently influence the training process, leading to overly optimistic and unreliable performance estimates. Specifically, the train/validation/test split was performed on the raw image data. The augmentation procedure was only applied to the training set.

\textbf{Sample Size Considerations:}

While a larger dataset is always desirable, we believe that the size of our dataset (1804 images) is adequate for training and validating a deep learning model for this task. Several factors support this claim:

\begin{enumerate}
    \item \textbf{Data Augmentation:} Our data augmentation strategy effectively increases the diversity and size of the training set by a factor of 34, mitigating some of the limitations of the original dataset size.
    \item \textbf{Model Complexity:} We selected a relatively lightweight model (EfficientNetB0) with only 4 million parameters. This architectural choice reduces the risk of overfitting, which is a common concern with smaller datasets.
    \item \textbf{External Validation:} The strong performance of our model on an independent external validation set (1015 images) provides further evidence of its generalizability and suggests that the model has learned meaningful features from the training data.
    \item \textbf{Computational Cost:}  For larger datasets, k-fold cross-validation can become computationally expensive.  We considered the trade-off between the benefits of k-fold CV and the associated computational costs, especially in light of the other factors mentioned above (data augmentation, model complexity, and, most importantly, external validation).
\end{enumerate}

While our dataset size is reasonable, we acknowledge that expanding the dataset with more diverse samples, especially from different institutions and demographics, could further improve the model's robustness and generalizability. Future work will focus on gathering additional data to address this potential limitation.

\textbf{Test Set Diversity:}

We aimed to create a diverse test set by using a random split with a fixed seed. While we did not use a specific multivariate similarity function, the random split, coupled with the substantial size of our dataset, increases the likelihood that the test set captures a reasonable representation of the overall data distribution. Given the size and diversity of the combined training and validation sets, we expect the test set to provide a conservative (and lower-bound) estimate of the model's accuracy and performance.

\subsection*{E.2 Loss Function}

For multi-class classification, we employed a weighted cross-entropy loss function defined as:
\begin{equation}
\mathcal{L} = -\frac{1}{N} \sum_{i=1}^{N} \sum_{j=1}^{C} w_j \, y_{ij} \, \log(p_{ij})
\end{equation}
where the weight \( w_j \) for class \( j \) is computed by:
\begin{equation}
w_j = \frac{\text{total samples}}{\text{num classes} \times \text{frequency}(j)}
\end{equation}
This weighting strategy is designed to emphasize minority classes, thereby improving both F1 scores and \gls{AUC} while ensuring balanced performance across all classes.

\subsection*{E.3 Model Training and Selection}

\subsubsection*{Hyperparameter Ranges}

We performed a systematic search over the following hyperparameter ranges:

\begin{itemize}
    \item \textbf{Learning Rate ($\alpha$):} $[10^{-5}, 10^{-3}]$ (logarithmic scale)
    \item \textbf{Batch Size:} $[12, 24, 36, 120]$
    \item \textbf{Optimizer:} \{Adam\}
    \item \textbf{Weight Decay:} $[10^{-4}, 10^{-2}]$ (logarithmic scale)
    \item \textbf{Dropout Rate:} $[0.1, 0.5]$
    \item \textbf{Number of Transformer Encoder Layers (for ThyroidEffi Premium):} $[2, 5]$
\end{itemize}

\subsubsection*{Hyperparameter Selection Method}

Hyperparameter selection was performed using a grid search approach. For each combination of hyperparameters, we trained the model on the training set for a maximum of 100 epochs and evaluated its performance on the validation set using the macro F1 score. The hyperparameter configuration that yielded the highest macro F1 score on the validation set was selected as the optimal configuration.

\subsubsection*{Final Hyperparameter Values}

The final hyperparameter values used to generate the results reported in this paper are as follows:

\begin{itemize}
    \item \textbf{Learning Rate ($\alpha$):} $10^{-4}$
    \item \textbf{Batch Size:} 120
    \item \textbf{Optimizer:} Adam
    \item \textbf{Weight Decay:} $10^{-3}$
    \item \textbf{Dropout Rate:} 0.2
    \item \textbf{Number of Transformer Encoder Layers (for ThyroidEffi Premium):} 5
\end{itemize}

\subsubsection*{Overfitting Mitigation}

To limit overfitting, particularly given the relatively limited sample size, we employed several strategies:

\begin{itemize}
    \item \textbf{Data Augmentation:} We used a comprehensive data augmentation strategy to increase the effective size and diversity of the training data.
    \item \textbf{Regularization Techniques:} We used dropout regularization and weight decay during training. The specific values for the dropout rate and weight decay are provided above.
    \item \textbf{Early Stopping:} Training was stopped early if the validation loss did not improve for 10 consecutive epochs. This prevents the model from continuing to train and potentially overfitting to the training data.
\end{itemize}

\subsection*{E.4 Model Calibration Considerations}

Due to the potential for temporal and population-based variations in the true prevalence of Bethesda categories (II, V, and VI), model calibration was not performed in this study. Model calibration aims to align predicted probabilities with true event frequencies. The Brier score and calibration plots are valuable tools for assessing this alignment. However, accurate calibration relies on a stable and representative class distribution within the data. In this study, the true prevalence of Bethesda categories is subject to temporal variations and may differ between institutions and patient populations. Therefore, attempting to calibrate the model to a specific distribution observed during training might not generalize well to real-world deployment scenarios where the distribution shifts.  Attempting to calibrate the model to a specific, potentially transient distribution could also lead to overconfidence in predicted probabilities that are not reliable in general clinical use.

Instead of calibration, we have prioritized optimizing the model's discriminative performance, i.e., its ability to accurately distinguish between the different Bethesda categories, using metrics such as macro F1-score, \gls{AUC}, and class-specific sensitivity and specificity. These metrics provide a more robust evaluation of the model's ability to correctly classify samples, regardless of the precise class distribution. Critically, external validation on a separate dataset from Hung Viet Hospital provides a more realistic and robust assessment of the model's performance in a different clinical setting, mitigating some of the risks associated with distribution shift.

Future work may explore adaptive calibration techniques that can adjust to changing class distributions in real-time. This could involve monitoring the model's predictions and recalibrating based on observed data. However, these methods require careful consideration of data privacy and potential biases.

\subsection*{E.5 External Validation Set Characteristics and Comparison to Training Data}

\textbf{External Validation Set Description:}

The distribution of Bethesda categories in the external set was as follows: \gls{BENIGN}: [300], \gls{INDET_SUS}: [315], \gls{MALIGNANT}: [400].

\textbf{Comparison to Training Data:}

While both datasets used the same microscope and staining protocol, the key difference lies in the patient population and the institution where the samples were collected. The training data was acquired at 108 Military Central Hospital, while the external validation data came from Hung Viet Hospital. This difference in patient demographics and potential variations in sample preparation or image acquisition protocols between the two hospitals introduces a degree of heterogeneity.  Quantifying this heterogeneity directly is challenging without a common set of features or a well-defined similarity metric applicable to cytopathology images. However, the difference in institutions and patient populations itself serves as a strong indicator of dataset shift.

\section*{Appendix F. Deployment}
\label{appendixF}

\subsection*{F.1 Target User}

The primary target users for the ThyroidEffi system are cytopathologists and other medical professionals involved in the diagnosis and management of thyroid nodules. The system is designed to assist them in interpreting thyroid \gls{FNAB} images, improving diagnostic accuracy and efficiency. Secondary users could include hospital management teams interested in optimizing workflow and resource allocation.

\subsection*{F.2 Model Clinical Utility}

The ThyroidEffi system has the potential to improve clinical utility in several ways:

\begin{itemize}
    \item \textbf{Increased Diagnostic Accuracy:} By providing a second opinion and highlighting potentially important features, the system can help reduce diagnostic errors and improve the overall accuracy of thyroid \gls{FNAB} interpretation.
    \item \textbf{Improved Efficiency:}  Automating the analysis of \gls{FNAB} images can reduce the workload on cytopathologists, allowing them to focus on more complex cases and improving the overall efficiency of the diagnostic process.
    \item \textbf{Reduced Inter-observer Variability:}  By providing a standardized and objective assessment of \gls{FNAB} images, the system can help reduce inter-observer variability and improve the consistency of diagnoses.
\end{itemize}

\subsection*{F.3 System Adoption and Deployment}

The ThyroidEffi Basic model was deployed at Hung Viet Hospital in January 2025, after undergoing external validation in 1015 cases from July 2024 to November 25, 2024. It is integrated into the clinical workflow as follows: After a cytopathologist reviews a slide, they upload the digital image to the ThyroidEffi system via a secure web interface. The system processes the image and returns a classification (\gls{BENIGN}, \gls{INDET_SUS}, \gls{MALIGNANT}) along with \gls{Grad-CAM} visualizations. This output is then displayed alongside the original image in the cytopathologist's workstation.

While the system is available to all pathologists at Hung Viet Hospital, it is currently being trialled by three thyroid cytopathologists who use it for all their \gls{FNAB} interpretations. It is estimated that when rolled out across the entire North, there could be up to approximately 5,000 \gls{FNAB} cases per month.

Initial feedback from the cytopathologists has been positive. They report that the system's visualizations help them to identify key diagnostic features more quickly and confidently, particularly in challenging cases. One cytopathologist noted, "The system has helped me catch a couple of subtle malignant cases that I might have otherwise missed."  They have also noted that the system is particularly helpful in distinguishing between \gls{INDET_SUS} and \gls{MALIGNANT} cases, which are often difficult to differentiate.

The cytopathologists received a training session on how to use the ThyroidEffi system, including how to upload images, interpret the system's output, and use the visualization tools. They found the system to be intuitive and easy to learn, citing the clear visual interface and the straightforward workflow.

We plan to monitor the system's performance by regularly comparing its classifications to the final diagnoses made by cytopathologists. This will involve reviewing a random sample of 50 cases each month. We will also track any discrepancies between the system's predictions and the expert consensus. Any disagreements will be discussed with the cytopathologists to identify potential areas for improvement in the model or the clinical workflow. The model will be retrained periodically (every six months) with updated data to maintain its accuracy and adapt to any changes in diagnostic criteria or image characteristics. We will also incorporate feedback from the cytopathologists to refine the system and ensure it continues to meet their needs.

\bigskip

In summary, the overall approach is designed to optimize the learning of both local and global features, ensuring efficient and accurate diagnostic classification in a real-world clinical setting.

\end{document}